\documentclass[preprint]{elsarticle}

\usepackage{amsmath,amssymb,amsfonts}
\usepackage{algorithmic}
\usepackage{graphicx}
\usepackage{textcomp}
\usepackage{xcolor}
\usepackage{booktabs}
\usepackage{enumitem}
\usepackage{mdframed}
\usepackage{pifont}% http://ctan.org/pkg/pifont

\def\BibTeX{{\rm B\kern-.05em{\sc i\kern-.025em b}\kern-.08em
    T\kern-.1667em\lower.7ex\hbox{E}\kern-.125emX}}

\usepackage{amsmath}
\usepackage{amssymb}
\usepackage{amsthm}
\usepackage{graphicx}
\usepackage{algorithm}
\biboptions{sort&compress}

\usepackage{tikz}

\def\halfcheckmark{\
\ding{51}\textsuperscript{\kern-0.5em\large\ding{55}}}

\renewcommand{\(}{\left(}
\renewcommand{\)}{\right)}

\newcommand{\n}{\boldsymbol{n}}

\newcommand{\s}{\mathbf{s}}

\newcommand{\Sig}{\boldsymbol{\Sigma}}

\newcommand{\z}{\mathbf{z}}
\newcommand{\w}{\mathbf{w}}

\newcommand{\E}{\mathbf{E}}

\newcommand{\0}{\mathbf{0}}

\newcommand{\1}{\mathbf{1}}

\newcommand{\x}{\boldsymbol{x}}

\renewcommand{\w}{\boldsymbol{w}}

\newcommand{\I}{{\boldsymbol{\mathcal{I}}}}

\newcommand{\EE}[1]{{\rm{E}}\left[#1\right]}

\newcommand{\norm}[1]{\left\|#1\right\|}

\renewcommand{\log}[1]{{\rm{log}}#1}
\renewcommand{\arg}[1]{{\rm{arg}}#1}

\newtheorem{lemma}{Lemma}
\newtheorem{definition}{Definition}
\newtheorem{theorem}{Theorem}

\newcommand{\Det} {\rm det}
\newcommand{\sumn}{\sum_{i=1}^N}
\newcommand{\prodn}{\prod_{i=1}^N}
\newcommand{\cmark}{\ding{51}}
\newcommand{\xmark}{\ding{55}}
\begin{document}
\begin{frontmatter}

\title{CFARnet: deep learning for target detection with constant false alarm rate}

% \author{\IEEEauthorblockN{1\textsuperscript{st} Tzvi  Diskin}
% \IEEEauthorblockA{\textit{School of Computer Science and Engineering} \\
% \textit{The Hebrew University of Jerusalem}\\
% Jerusalem, Israel \\
% tzvi.diskin@mail.huji.ac.il}
% \and
% \IEEEauthorblockN{2\textsuperscript{nd} Uri  Okun}
% \IEEEauthorblockA{
% uri.okun@gmail.com}
% \and
% \IEEEauthorblockN{3\textsuperscript{rd} Ami  Wiesel}
% \IEEEauthorblockA{\textit{School of Computer Science and Engineering} \\
% \textit{The Hebrew University of Jerusalem}\\
% Jerusalem, Israel \\
% ami.wiesel@mail.huji.ac.il}}

\author{Tzvi Diskin, Yiftach Beer, Uri Okun and Ami Wiesel}
\date{July 2022}

\begin{abstract}
We consider the problem of target detection with a constant false alarm rate (CFAR). This constraint is crucial in many practical applications and is a standard requirement in classical composite hypothesis testing.
In settings where classical approaches are computationally expensive or where only data samples are given, machine learning methodologies are advantageous. CFAR is less understood in these settings. To close this gap, we introduce a framework of CFAR constrained detectors. Theoretically, we prove that a CFAR constrained Bayes optimal detector is asymptotically equivalent to the classical generalized likelihood ratio test (GLRT). Practically, we develop a deep learning framework for fitting neural networks that approximate it. Experiments of target detection in different setting demonstrate that the proposed CFARnet allows a flexible tradeoff between CFAR and accuracy. 

\end{abstract}

\end{frontmatter}

\section{Introduction}

%\textcolor{teal}{Our story is "1. in the past, there were classical solutions which had guaranteed properties but they were not optimal and they were not always tractable. 2. Then deep learning arrived, solving both a. and b. , but we lost the guaranteed properties and it's a big deal. 3. In this paper we get it back.". But the introduction starts from 2. Can you add 1? I'd write it instead of "The deep learning revolution [...] detection [7-12]", and perhaps place the citations after the "are intractable". Your (Tzvi/Ami) phrasing here will probably be much better than mine.} 

The deep learning revolution has led many to apply machine learning methods to classical problems in all fields. Examples range from estimation \cite{dong2015image, ongie2020deep, gabrielli2017introducing, dua2011artificial,kerbaa2023multi, dreifuerst2021signalnet,diskin2021learning} to detection \cite{samuel2019learning,girard2021deep,brighente2019machine,de2017approximating,ziemann2018machine,theiler2021bayesian, addabbo2023application}. Deep learning is a promising approach for developing high accuracy and low complexity alternatives when classical solutions are intractable. However, deep learning based methods lack some of the crucial guarantees that classical methods provide.   In this paper, we focus on learning detectors for composite hypothesis testing. We show that current solutions lack the Constant False Alarm Rate (CFAR) requirement which allows robust performance under different conditions and is critical in many applications. To close this gap, we provide a framework for learning accurate CFAR detectors. 

%Hypothesis testing and target detection are fundamental problems in statistical signal processing. Applications range from radar and imaging to communication systems. 

%Traditionally, hypothesis testing and target detection were solved using tools that choose the hypothesis which is more likely. These model-driven tools led to high accuracy but were often computationally intensive. On the other hand, machine learning considers low-cost classifiers that are fitted to minimize data-driven loss functions. Modern testing problems are somewhere in the middle: it is common to assume a partially-specified and physically-based statistical model, and there are often computational constraints on the final detector. Therefore there is a natural trend to switch to learned detectors. These usually deliver on their promise, but lack the CFAR property which is critical in many applications. 

%In the first part of this paper, we consider the use of machine learning for simple hypothesis testing and review state of the art. The main conclusion is that machine learning provides a promising alternative to classical methods. In the second part, we move on to composite hypothesis testing with unknown parameters. In these more challenging problems we claim that existing methods are insufficient and propose alternative solutions that provide the best of both worlds. 

Detection theory begins with simple hypothesis testing where a detector needs to decide between two fully specified distributions. The classical solution is the Likelihood Ratio Test (LRT) which is optimal in terms of maximizing the detection probability subject to a false alarm constraint. Recently, there is a growing trend of switching from model-based detectors to data-driven classifiers.
%In deep learning based detectors, instead of using the statistical model directly, the goal is minimize the average error on a training dataset among a restricted class of classifiers. 
In simple settings, it is well known that the optimal Bayes classifier converges to the LRT with a specific false alarm rate \cite{kay1998fundamentals,bartlett2006convexity}. Learned classifiers can be interpreted as approximations to these optimal Bayes solutions.
More advanced classifiers can also maximize the cumulative detection rate over a wide range of false alarms, also known as the (partial) area under the curve (AUC) \cite{herschtal2004optimising, brefeld2005auc, narasimhan2013structural}. Large deviation analysis is available in  \cite{braca2022statistical}.
Examples of works on machine learning for target detection include \cite{ziemann2018machine,girard2021deep}  in hyperspectral imagery and \cite{de2017approximating} in radar.

Composite hypothesis testing is a more challenging setting where the hypotheses involve unknown deterministic parameters. A CFAR detector is invariant to these parameters and has constant false alarm probabilities. The CFAR property is important in many applications. It guarantees consistent and predictable performance across different environments. Indeed, critical systems often need to satisfy false alarm requirements in multiple environments simultaneously. CFAR allows the user to set the thresholds independently of the specific environment. %Thus in non CFAR detectors, thresholds must be set to accommodate the worst-case scenario, leading to a decrease of the detection probability on the other environments. 
More generally, the CFAR property is related to predictive equality and equal opportunity properties that play an important role in the field of fair machine learning \cite{verma2018fairness}.

Model based detectors usually satisfy the CFAR property. The most popular approach is the Generalized Likelihood Ratio Test (GLRT) which is defined by first estimating the unknown parameters and then plugging them into a standard LRT. Among its other favorable properties, GLRT is known to be asymptotically CFAR. % and performs well when the model is exact and the number of unknowns is relatively small. Otherwise, it is generally sub-optimal and may be computationally expensive.
Alternatively, there are many works on designing CFAR detectors for specific families of distributions \cite{gini2002covariance, conte2003cfar,kokaly2017usgs,coluccia2022design, coluccia2022glrt}. In paricular, CFAR detectors for radar processing were developed by relying on CFAR features \cite{lin2019dl,akhtar2018neural,akhtar2021training}. 

%In composite hypothesis testing, the picture is more complicated as unlike the equivalence of LRT with the optimal Bayes detectors, there is no such equivalence between the classical and the Bayesian detectors. 
%An important ingredient of model-based machine learning is synthetic generation of a training dataset. In simple hypothesis settings, it is straight forward to generate artificial samples for the two hypotheses and train a network to classify them. The process is more involved in composite hypothesis testing where additional unknown parameters without any prior must be generated. In many real world applications, it is common to rely on a hybrid training set where real noise or clutter data is augmented with synthetically planted targets, e.g., \cite{ziemann2018machine}. In all of these a key theoretical question is the robustness of the learned network to deviations from the synthetic assumptions.  

The main contributions of the paper are:
\begin{itemize}
    \item We define a general framework for Bayesian and learning-based CFAR detectors that can be applied to arbitrary composite hypothesis testing.
    \item We analyze the asymptotic performance of the proposed detectors. We prove that the CFAR constrained Bayes detector is asymptotically equivalent to the popular GLRT.
    \item We develop CFARnet - a practical deep learning approach to fitting neural networks with a CFAR constraint. To optimize CFARnet, we rely on empirical and differentiable distances that have recently become popular in unsupervised deep learning \cite{goodfellow2014generative,li2015generative,gretton2012kernel}. 
    \item We demonstrate the advantages of CFARnet on classical detection problems. In asymptotic settings where GLRT performs well, CFARnet achieves the same performance with lower computational complexity. In non-asymptotic settings where GLRT is suboptimal, CFARnet can outperform it and still guarantee near-CFAR behavior. Finally, experiments show that CFARnet is often also preferable in terms of performance on the worst case environment. 
\end{itemize}

The paper is organized as follows: In Section 2, we formalize
the composite hypothesis testing problem. Next,
in Section 3 we introduce the CFAR constrained Bayes detector and prove its asymptotic equivalence to GLRT. In section 4, we define CFARnet as an approximation of the CFAR constrained Bayes detector. We implement CFARnet and demonstrate its performance on different settings in section 5. Finally, we
conclude and discuss some limitations of the work in Section
6.

%Fair learning tries to eliminate biases in the training set and considers properties that need to be protected \cite{agarwal2019fair,bagnell2005robust}. OOD works introduce an additional ``environment'' variable and the goal is to train a model that will generalize well on new unseen environments \cite{creager2021environment,maity2020there,arjovsky2019invariant,wald2021calibration}. 

%Outline ... Notations...

%Another downside to LRT is when the likelihoods are not a priori known and must be estimated from data. In this case, alternative solutions may be advantageous. In this paper, we focus on signal processing applications where an exact physics based model is known. In Section ?, we will also consider partially known models. Fully data-driven hypothesis testing is outside the scope of this paper. 

\section{Problem formulation}
We consider a binary hypothesis test. Let $\x$ be an observed random vector whose distribution $p(\x;\z)$ depends on an unknown deterministic parameter $\z$. The value of $\z$ defines two possible hypotheses
\begin{align}\label{comp_testing}
    &y=0:\quad \z\in{\mathcal{Z}}_0\nonumber\\
    &y=1:\quad \z\in{\mathcal{Z}}_1.
\end{align}
It is customary to divide $\z=\{\z_r,\z_n\}$ into two components. The parameter $\z_r$ is discriminative whereas  $\z_n$ is a nuisance parameter which is the same under both hypotheses.

Throughout the paper, we will illustrate the ideas using a simple and classical running example:

\begin{mdframed}
{\bf{Running example:}} Target detection when both the target amplitude and the noise scaling are unknown:
\begin{align}\label{DC}
    \x=A\1+\sigma\n
\end{align}
where $\1$ is a target vector of ones, $\n$ is a random vector with independent and identically distributed (i.i.d.) ${\mathcal{N}}(0,1)$ noise variables, $\z=[A,\sigma]$ are deterministic unknown parameters, and 
\begin{align}\label{params}
    &{\mathcal{Z}}_0=\left\{\z\;:\;A=0,\; \sigma\neq 0\right\}\nonumber\\
    &{\mathcal{Z}}_1=\left\{\z\;:\;A\neq 0,\; \sigma\neq 0\right\}
\end{align}
Here, $A$ acts as a discriminative parameter that varies between the two hypotheses, whereas the noise variance $\sigma^2$ is a nuisance parameter which is identical in both. Throughout the paper, we will consider both cases when $\sigma^2$ is known and unknown.  %A classical example is a signal with an unknown amplitude in a noise with unknown variance. The amplitude is a discriminative parameter because it is zero in the $y=0$ hypothesis and non-zero in the $y=1$ hypothesis. The variance of the noise is considered a nuisance parameter as it does not depend on whether the signal exists or not. \anote{Ami - please check}
\end{mdframed}

The goal is to design a detector $\hat y(\x)\in\{0,1\}$ as a function of $\x$ that will identify the true hypothesis $y\in\{0,1\}$. 
Performance is measured in terms of probability of correct detection, also known as True Positive Rate (TPR):
\begin{align}\label{PD_FA1}
   {\rm Pr}_{\rm TPR}(\z) =  \; &{\rm Pr}(\hat y(\x) =1 ;y=1)
\end{align}
and probability of false alarm, also known as False Positive Rate (FPR):
\begin{align}\label{PD_FA2}
   {\rm Pr}_{\rm FPR}(\z) =  {\rm Pr}(\hat y(\x) =1 ;y=0).
\end{align}
In practice, the user typically provides a false alarm constraint ${\rm Pr}_{\rm FPR}\leq \alpha$ that must be satisfied and the goal is to maximize ${\rm Pr}_{\rm TPR}$.

It is standard to consider decision functions of the form
\begin{equation}\label{detector}
    \hat{y}\left(\x\right)=\1_{T(\x)\geq\gamma}=\begin{cases}
0 & T\left(\x\right)<\gamma\\
1 & T\left(\x\right)\geq\gamma
\end{cases}, 
\end{equation}
where $T(\x)$ is denoted as the {\textbf{detector}} function and $\gamma$ is a {\textbf{threshold}} value. This structure allows users to tune the FPR by adjusting the threshold.  Performance is usually visualized using the Receiver Operating Characteristic (ROC) which plots the TPR as a function of the FPR. In signal processing applications, users are often interested in a region of very low FPRs, e.g., $10^{-1}-10^{-3}$ and the goal is to maximize the TPR probabilities in this area. Note that the ROC does not give a full specification of the detector as it assumes that the threshold is tuned to fit the level of the FPR in each point of the curve. This leads us to a main challenge in detection theory, namely the unknown nuisance  parameters under the null hypothesis $y=0$, e.g., the unknown noise variance $\sigma^2$ in the running example. The FPR is generally a function of these parameters and cannot be controlled without their knowledge. As a remedy it is often preferable to restrict the attention to CFAR detectors.
\begin{definition}
A detector $T(\x)$ is CFAR if its FPR ${\rm Pr}\left(T(x)>\gamma|\z\in\mathcal{Z}_0\right)$ is invariant
to the value of $\z\in{\mathcal{Z}}_0$, for any threshold $\gamma$.
\end{definition}
It is straightforward to see that the above definition is equivalent to invariance of the distribution of $T(\x)$ to all $\z\in{\mathcal{Z}}_0$. %, as ${\rm Pr}\left(T(x)>\gamma|\z\in\mathcal{Z}_0\right)=1-F_T(\gamma|\z\in\mathcal{Z}_0)$ where $F_T$ is the cumulative distribution function (CDF) of the detector. Thus a detector is CFAR if and only if its distribution is invariant to all $\z\in{\mathcal{Z}}_0$. 

%In Example 1, $T(\x)$ is a CFAR test if its distribution is invariant to the value of $\sigma^2$. A natural detector for this example is the naive test $(\x^T\1)^2$ which performs very well in terms of ROC but is not CFAR. 

As we will review below, many classical detectors are CFAR or asymptotically CFAR. With the growing trend of switching to machine learning, the goal of this paper is to introduce a competing framework for learning CFAR detectors.

\section{Model based detectors}
\subsection{Classical Detectors}
Traditionally, detectors were developed based on statistical models using likelihood ratios. In the simple case, all the parameters of the hypotheses are known (e.g., the running example if $\sigma$ was known and $A$ and had a single possible value under $y=1$). In this case, hypothesis testing has an optimal solution known as the Likelihood Ratio Test (LRT) due to Neyman-Pearson lemma \cite[p. 65]{kay1998fundamentals}: 
%LRT theory states that the optimal detector for maximizing detection subject to a given false alarm probability is:
\begin{align}\label{LRT}
    T_{\rm LRT}(\x) &=2\log\frac{p(\x;\z=\z_1)}{p(\x;\z=\z_0)},
\end{align}
where the threshold $\gamma$ is chosen to satisfy the false alarm (FPR) constraint.

The more realistic scenario is composite hypotheses testing where one or both of the hypotheses allow multiple possible values and there is no solution that is optimal for all of them simultaneously. A popular heuristic is the Generalized Likelihood Ratio Test (GLRT) that estimates the unknowns using the Maximum Likelihood (ML) technique and plugs them into the LRT detector \cite[p. 200]{kay1998fundamentals}:
\begin{align}\label{GLRT}
    T_{\rm GLRT}(\x) &=  2\log\frac{\max_{\z\in{\mathcal{Z}}_1}p(\x;\z)}{\max_{\z\in{\mathcal{Z}}_0}p(\x;\z)}.
\end{align}
Setting the threshold to ensure a fixed ${\rm Pr}_{\rm{FPR}}$ is not trivial. Fortunately, under regularity conditions, GLRT is asymptotically CFAR and its threshold can be set for all values of the unknown parameters simultaneously \cite[p. 206]{kay1998fundamentals}. 

\begin{mdframed}
{\bf{Running Example:}} If the level of the noise $\sigma^2$ is known (no nuisance parameters), then GLRT is simply
\begin{align}\label{glrt_known}
    T_{\rm GLRT}(\x) &= \frac{\(\x^T \1\)^2}{N\sigma^2}.
\end{align}
When $\sigma^2$ is unknown, (\ref{glrt_known}) can still be used without the constant denominator. This will result in the same ROC performance but no CFAR. A better approach is the GLRT associated with an unknown $\sigma^2$ which can be derived as
\begin{align}\label{GLRT uncorelated}
    T_{\rm{GLRT}}(\x)=\frac{(\x^T\1)^2}{\x^T\x}.
\end{align}
This GLRT guarantees both a similar ROC performance and a CFAR for any value of $\sigma^2$. Indeed, the denominator of (\ref{GLRT uncorelated}) can be interpreted as an estimator of the unknown variance, i.e., $\x^T\x\approx N\sigma^2$.
\end{mdframed}

% \begin{align}\label{fim}
%     &\lambda=(\z_{r1}-\z_{r0})^T\F(\z_{r1}-\z_{r0})\nonumber\\
%     &\F = \I_{\z_r\r_r}-\I_{\z_r\r_s}\I^{-1}_{\z_s\r_s}\I_{\z_s\r_r}
% \end{align}
% where $\z_{r1}$ is the ground truth parameter and the $\I$ matrices denote the different sub-blocks of the  Fisher Information Matrix (FIM) as detailed in \cite[6.5]{kay1998fundamentals}. Thus, GLRT is asymptotically CFAR and its threshold can be set using (\ref{asympP}).

%This leads to a closed form threshold which guarantees an asymptotic constant false alarm rate (CFAR) for any nuisance parameter $\z_s$.

GLRT is probably the most popular solution to composite hypothesis testing. It gives a simple recipe that performs well under asymptotic conditions. Its main downsides are that it is sensitive to deviations from its theoretical model, it is generally sub-optimal under finite sample settings and that it may be computationally expensive as both the numerator and denominator of the GLRT involve optimization problems that may be large scale, non-linear and non-convex. Therefore, there is an ongoing search for flexible, robust and low cost alternatives. 

\subsection{Bayesian detectors}
In this section, we review the Bayesian approach to hypothesis testing \cite[Sec. 6.4.1]{kay1998fundamentals}. As expected, this approach does not lead to CFAR detectors. To close this gap, we introduce a CFAR constrained Bayesian detector. We then analyze the detectors under the classical large data record setting.

A competing approach to hypothesis testing is based on the Bayesian methodology. The latter differs in two (related) aspects from the classical approach. First, the unknown parameters ($y$ and $\z$) are random with known priors ${\rm Pr}(y)$ and $p(\z|y)$. As detailed in \cite[Sec. 6.4.1]{kay1998fundamentals} the choice of these priors is often difficult, and practitioners resort to fictitious ``flat'' distributions which are non informative. 
Second, using these priors, the joint distribution $p(\x,y,\z)$ can be used to express a single measure of the error (the Bayes risk). A popular Bayes risk is the probability of error:
\begin{align}\label{bayes risk}
   {\rm Pr}_{\rm ERR}={\rm Pr}(\hat y \neq y)\ =  \sum_{y=0}^1\int \1_{\hat{y}\neq y}p(\x,y,\z)d\x dy d\z ,
\end{align}

Minimizing (\ref{bayes risk}) leads to a well defined Bayes optimal detector which is also known as Bayesian LRT (BLRT):
\begin{align}\label{ Bayes optimal}
    {\rm BLRT:}\quad \min_{T,\gamma}  {\rm Pr}_{\rm ERR}( T,\gamma).
\end{align}
Its solution is
\begin{align}\label{BLRT}
    &T_{\rm BLRT}(\x)=2\log\frac{p_1(\x)}{p_0(\x)}=2\log\frac{\int_{\z\in\mathcal{Z}_1} p(\x;\z)p(\z)d\z}{\int_{\z\in\mathcal{Z}_0}p(\x;\z)p(\z)d\z}\nonumber\\
    &\gamma_{\rm BLRT} = 2 {\rm log}\frac{{\rm Pr}(y=0)}{{\rm Pr}(y=1)}.
\end{align}

Originally, BLRT was designed to minimize the probability of error. However, practitioners often use it even when the underlying formulation is classical, by assigning fictitious priors to the unknown parameters. In simple hypotheses,  BLRT is identical to LRT and the only difference is the thresholds. For any required FPR, BLRT with an appropriate threshold maximizes the TPR (independently of the chosen prior). 

%\subsection{Composite hypotheses testing}
In the composite case, BLRT is less understood. Due to the integrals, it typically does not have an easy solution. Even in the simple running example, BLRT may have a complicated form that depends on the chosen priors. Moreover, it is not clear how it performs in terms of FPR and TPR, nor how it compares to GLRT. Experiments in different settings reveal that BLRT does not generally guarantee a CFAR.
\subsection{Our Proposal - CFAR Bayesian detector}
To close this gap, we introduce a new detector, called CLRT, which is Bayes optimal subject to a CFAR constraint: 
\begin{align}\label{CFAR optimization}
    {\rm CLRT:}\quad \bigg\{\begin{array}{ll}
        \min_{{T},\gamma} & {\rm Pr}_{\rm ERR}(T,\gamma) \\
        {\rm s.t.} & {T} \; {\rm is\;CFAR}
    \end{array}.
\end{align}
%It can be shown that CLRT is asymptotically equivalent to GLRT.

 In what follows, we claim that CLRT is the natural Bayesian version of GLRT. When there are no complicating nuisance parameters, BLRT and GLRT are asymptotically equivalent. Otherwise, the CFAR constraint in (\ref{CFAR optimization}) leads to equivalence between CLRT and GLRT. To formally state this result, we begin by recalling the classical large data records setup as detailed in  \cite[p. 205]{kay1998fundamentals}. We consider $\x\sim p(\x;\z_r,\z_n)$ and test 
\begin{align}\label{comp_testing2}
    &y=0:\quad \z_r=\z_{r_0},\z_n \nonumber \\
    &y=1:\quad \z_r\neq \z_{r_0}, \z_n, 
\end{align}
where $\z_r\in\mathbb{R}^{d_r}$ is a discriminative parameter and  $\z_n\in\mathbb{R}^{d_n}$ is a nuisance parameter. We let $\z_{r_0}$ and $\z_{r_1}$ be the true values of $\z_r$ at $y=0$ and $y=1$, respectively. We further assume that:
\begin{itemize}
    \item The data consist of many i.i.d samples from the true statistical model: 
    \begin{align}
p(\x;\z_r,\z_n)=\prod_{i=1}^Np(x_i;\z_r,\z_n), \qquad N\rightarrow\infty
    \end{align}
    \item The signal is weak:
    \begin{align}
     \norm{\z_{r_1}-\z_{r_0}}=\frac{s}{\sqrt{N}}, 
    \end{align}
    where $s$ is some finite constant. 
    \item The ML estimators of the unknown parameters are statistically efficient and attain their asymptotic performance.
\end{itemize}
We note here that the weak signal assumption ensures that the performance remains independent of $N$, when $N\rightarrow\infty$. Practically, it represents a problem of a weak signal such that a large the number of measurements are needed in order to detect it with a reasonable accuracy.
Under these conditions, when $N\rightarrow \infty$, it is well known that GLRT attains its asymptotic performance. The next two theorems analyze BLRT and CLRT in the same setting. We first state the results and their consequences, and then provide the proofs. %Before we present the theorem, we remind that two tests are equivalent if they are equal up to a \textbf{known} monotonic function that is applied on both the detector and the threshold.
\begin{theorem}\label{no nuisance}
    Consider the classical asymptotic setting and assume the technical conditions as detailed in the appendix. Then, independently of the choice of $p(\z)$ prior, we have
    \begin{align}\label{BLRT Laplace}
    T_{\rm BLRT}(\x) &\rightarrow T_{\rm GLRT}(\x)+{\rm func}(\z_{r_0},\z_n).
\end{align}
\end{theorem}
Note that the function ${\rm func(\z_{r_0},\z_n)}$ does not depend on $\x$ and that $\z_{r_0}$ is known. Therefore, BLRT and GLRT are equivalent if there are no nuisance parameters (no $\z_n$). Otherwise, the performance of BLRT depends on the value of the unknown $\z_n$ parameter, whereas GLRT is CFAR. Thus, the detectors are not generally equivalent as the next example demonstrates.

\begin{mdframed}
{\bf{Running Example:}} To derive the Bayesian versions of the detectors, we assume a fictitious prior $A\sim {\mathcal{N}}(0,\sigma_r^2)$. If the  noise variance $\sigma^2$ is known (no nuisance parameters), then BLRT is given by
\begin{align}
T_{\rm BLRT}(\x) &= \frac{\(\x^T \1\)^2}{N\sigma^2+\sigma_r^2} +\log\(\frac{\sigma^2}{\sigma^2+N\sigma_r^2}\)
\end{align}
The variances are all known and appropriate thresholds can be selected so that BLRT and GLRT in (\ref{glrt_known}) are asymptotically equivalent.
On the other hand, if $\sigma^2$ unknown then BLRT is more complicated and does not satisfy a simple closed form solution. Yet, independently of the chosen prior for $\sigma^2$, its limit can be derived as
\begin{align}
    T_{\rm BLRT}(\x) &\stackrel{N\rightarrow\infty}{\rightarrow} \frac{\(\x^T \1\)^2}{\x^T\x} +\log\(\frac{\x^T\x}{N^2\sigma_r^2}\).
\end{align}
Evidently, this asymptotic BLRT and the GLRT in (\ref{GLRT uncorelated}) are not equivalent. Their ROC performance are similar as $\x^T\x\rightarrow N\sigma^2$ but BLRT requires different thresholds and does not have a CFAR.     
\end{mdframed}

Interestingly, the next theorem shows that CLRT, with the added CFAR constraint, is equivalent to GLRT.

\begin{theorem}
    Consider the classical asymptotic setting with a block diagonal Fisher Information Matrix (FIM) and the technical assumptions  detailed in the appendix. Then, independently of the choice of $p(\z)$ prior, CLRT is equivalent to GLRT, that is, GLRT is a solution to (\ref{CFAR optimization}).
\end{theorem}

\begin{mdframed}
{\bf{Running Example:}} Asymptotically, the GLRT in (\ref{GLRT uncorelated}) is a solution to CLRT in (\ref{CFAR optimization}).
\end{mdframed}

\section{Learned detectors}\label{ml4lrt}
In this section, we propose a framework for learning neural networks that approximate the theoretical CLRT detector.
\subsection{Background and existing work}
Deep learning is based on minimizing the empirical error on a training data set, among a class of detectors that is parameterized by a deep neural network (DNN) architecture. Deep learning is usually applied on problems where the statistical model is unknown but a large dataset of labeled examples is provided. A different scenario is when the statistical model is known, but classical detectors are intractable or computationally expansive. In such cases, a DNN can be trained on synthetic data that is generated using the physical model \cite{samuel2019learning}. In such setups, the DNN is simply an approximation of the best Bayesian detector, where the true Bayes risk is replaced by the empirical error. It relies on a computationally intensive fitting
phase which is done offline, and yields a DNN
with fixed complexity that can be easily applied in inference
time.

We give now a short description of this process.
The dataset is generated using the (possibly fictitious) priors  ${\rm Pr}(y)$, $p(\z;y)$ and the probabilistic model $p(\x;\z)$. That is, first $y$ and $\z$ are generated according to their priors. For each $y^{(k)}$ and $\z^{(k)}$, a measurement $\x^{(k)}$ is generated according to the true $p\left(\boldsymbol{x};{\z^{(k)}}\right)$. Together, we obtain a synthetic dataset:
\begin{equation}
    {\mathcal{D}}_N=\{\boldsymbol{x}^{(k)},\z^{(k)},y^{(k)}\}_{k=1}^K.
\end{equation}

Next, a class of possible detectors ${\mathcal{T}}$ is chosen in order to tradeoff expressive power with computational complexity in test time. The class is usually a fixed differentiable neural network architecture. In our context, it also makes sense to reuse existing ingredients from classical detector as non-linear features or internal sub-blocks \cite{akhtar2021training, samuel2019learning}.

Finally, the learned detector is defined as the minimizer of an empirical loss function
\begin{equation}\label{classifcation}
   \min_{\hat{{T}}\in {\mathcal{T}}} \frac 1K \sum_{k=1}^K L(\hat T(\x^{(k)}),y^{(k)}).
\end{equation}
where $L(\cdot;\cdot)$ is a classification loss function. Ideally, we would like to minimize the zero-one loss which corresponds to the average probability of error. Practically, for efficient optimization, a smooth and convex surrogate loss, as the hinge or cross entropy functions, is minimized by stocahstic gradient decent (SGD) and its extensions \cite{shalev2014understanding}. The overall procedure for learning a detector is summarized in Algorithm 1.

\begin{algorithm}
\caption{Bnet: learning an approximation to BLRT.}\label{alg:BCE}
\begin{itemize}
    \item Require   $p(\x;\z)$.
    \item Choose ${\rm Pr}(y)$ and $p(\z;y)$.
    \item For each $k=1,\cdots,K$:
    
    \hspace{4mm}Generate $y^{(k)}$.
    
    \hspace{4mm}Generate $\z^{(k)}$ given $y^{(k)}$.
    
    \hspace{4mm}Generate $\x^{(k)}$ given $\z^{(k)}$.
    \item Solve 
    
    \hspace{4mm}$\min_{\hat{{T}}\in {\mathcal{T}}} \frac 1K \sum_{k=1}^K L(\hat T(\x^{(k)}),y^{(k)})$.
\end{itemize}
\end{algorithm}

%Practically, the minimization problem is usually solved by stocahstic gradient decent (SGD) and its extensions \cite{shalev2014understanding}. Learned detectors can efficiently approximate any required BLRT. This requires a large enough training dataset, an expressive enough class of possible detectors $\mathcal{T}$ and a consistent surrogates losses of the probability of error, as the hinge loss \cite{bartlett2006convexity}. When the hypotheses are simple, such solutions are also optimal in the classical sense.

\subsection{CFARnet}
We now show that CLRT can also be approximated using a DNN denoted by CFARnet. Like Bnet, the approximation will be accurate if the training dataset is large enough and the class of detectors is expressive enough. The only difference in CFARnet is the introduction of an additional differentiable CFAR penalty which approximates the constraint. The rest of this subsection provides its details. 

To approximate the CFAR constraint
 CFARnet introduces two modifications to Algorithm 1. First, we augment the classification loss with a penalty function that ensures similar distributions of the detector $\hat T(\x)$ for all values of $\z\in {\mathcal{H}}_0$. Second, in order to compare such distributions empirically, we rely on an enhanced training set that includes multiple $\{\x^{(k,m)}\}_{m=1}^M$ for each $\z^{(k)}$.

% \begin{equation}\label{dpd}
%   \min_{\hat{{T}}\in {\mathcal{H}}} \frac 1N \sum_{i=1}^N L(\hat T(\x_i),\z_i)+ \alpha R_{\rm{CFAR}}(\hat{T})
% \end{equation}where $\alpha$ is a regularization hyper-parameter. The regularizer 

The main idea is adding a penalty to the objective function that promotes a CFAR detector. For this purpose, we need to measure the distance between different distributions.
\begin{definition}
Let $X\sim p(X)$ and $Y\sim p(Y)$ be two random variables. A statistical distance $d(X;Y)$ is a function
that satisfies $d(X;Y)\geq 0$ with equality if and only if $p(X)=p(Y)$. In particular, a statistical distance $d(X;Y)$ can be empirically estimated using a set of data realizations $\hat d(\{X^{(m)}\}_{m=1}^M,\{Y^{(m)}\}_{m=1}^M)$.
\end{definition}
Given a statistical distance $d(\cdot;\cdot)$, the penalty is defined as a sum of distances between the distributions of $\hat{T}$ under different values of $\z$:
\begin{align}\label{const2}
    R(\hat{T})=\sum_{\z, \z'\in {\mathcal{Z}}_0} d\left(\hat{T}(\x); \hat{T}(\x')\right)
\end{align}
where
\begin{align}
    &\x\sim p(\x;\z)\nonumber\\
    &\x'\sim p(\x;\z').
\end{align}
Clearly, any CFAR test must satisfy $R(\hat{T})=0$. 
A similar approach can be found in \cite{romano2020achieving} which  enforces ``equalized odds'' using a distance between distributions. A main difference is that CFAR is a one-sided fairness property and requires equal rates only in the null hypothesis. Algorithmically, \cite{romano2020achieving} compares the high dimensional joint distribution of the predictions and the unknown parameters, whereas we only consider the scalar distribution of the predictions. This makes our method significantly cheaper in terms of computational complexity.

Practically, to minimize (\ref{const2}), we use empirical estimates of the distances where each distribution is represented using a small dataset. For each $\z^{(k)}$, we synthetically generate multiple observations $\x^{(k,m)}$ for $m=1,\cdots,M$. Similarly, for each $\z^{(k')}$ we compute multiple $\x^{(k',m)}$. We then plug these into the empirical distances:

\begin{align}\label{const22}
    \hat R(\hat T)=\sum_{\tiny{\begin{array}{ll}
         k: y^{(k)}=0  \\
         k': y^{(k')}=0 
    \end{array}
    }} \hat d\left(\{\hat{T}(\x^{(k,m)}\}_{m=1}^M); \{\hat{T}(\x^{(k',m)})\}_{m=1}^M\right)
\end{align}
where the sum is only computed with respect to samples corresponding to $y=0$.

Our implementation of CFARnet uses the Maximum Mean Discrepancy (MMD) distance \cite{gretton2012kernel} as detailed in \ref{app_dist}. We also use a hyper-parameter $\lambda>0$ that trades off the importance of the classification accuracy versus the CFAR penalty. The overall CFARnet procedure is summarized in Algorithm 2. If $N$, $M$ and $\lambda$ are large enough and the architecture is expressive enough then the global solution of this algorithm is a good approximation of CRLT.

\begin{algorithm}
\caption{CFARnet: learning an approximation to CLRT}\label{alg:cfar}
\begin{itemize}
    \item Require $p(\x;\z)$.
    \item Choose ${\rm Pr}(y)$ and  $p(\z;y)$.
    \item For each $k=1,\cdots,K$:
    
    \hspace{4mm}Generate $y^{(k)}$.
    
    \hspace{4mm}Generate $\z^{(k)}$ given $y^{(k)}$.
    
    \hspace{4mm}For $m=1,\cdots,M$:
    
    \hspace{8mm} Generate $\x^{(k,m)}$ given $\z^{(k)}$.
    \item Solve 
    
    \hspace{4mm}$\min_{\hat{{T}}\in {\mathcal{T}}} \frac 1{KM} \sum_{k,m} L(\hat T(\x^{(k,m)},y^{(k)})+\lambda \hat R(\hat T)$.
\end{itemize}
\end{algorithm}
 %As in the case of Algorithm 1, here also the minimization problem can be solved using SGD. It easy to see that when $M,N\rightarrow\infty$, for $\lambda\rightarrow\infty$, a CFAR detector have smaller augmented loss then any other detector. 
 %Together with the classification loss, it will results with the most accurate (with a respect to the selected priors and loss function) among all the CFAR detectors. As will shown in the next section, practically the parameter $\lambda$ will be finite, leading to a near CFAR detector.

%The theorems give more insight on CFARnet as a deep learning approximation of BayesCFAR. In the asymptotic setting with block diagonal FIM, GLRT is CFAR and there is no degradation in ROC performance due to the additional constraint. More generally, there is an inherent tradeoff between accuracy and CFAR and a sweet spot must be identified for each application. The theorems also gives a new interpretation to the GLRT, Wald  and Rao tests, that have the same asymptotic distributions \cite[p. 205]{kay1998fundamentals} but do not directly minimize any cost function. Apparently, in asymptotic setting with block diagonal FIM, they are the minimizers of the Bayesian 0-1 loss subject to the CFAR constraint.

\section{Numerical experiments}
In this section, we provide results of numerical experiments that illustrate the different properties and advantages of CFARnet. In each experiment, we compare 3-4 detectors: one or two classical baselines (GLRT and its variants), a non-CFAR Bnet neural network and our proposed CFARnet which is identical to Bnet but is trained with a CFAR loss as discussed above. We measure performance using four criteria:
\begin{itemize}
    \item ROC: ROC area for fixed nuisance parameters.
    \item CFAR: difference between FPRs at different values of the nuisance parameters.
    \item NP: Neyman-Pearson performance defined as TPR for a strict FPR $\leq 0.01$ constraint for all nuisance parameters. TPR is plotted as a function of the nuisance parameter.
    \item SPEED: computational complexity of the detector at inference time. SPEED is measured defined as the inference time on a test set of 10000 samples, in milliseconds (performed on an Nvidia T4 GPU).
\end{itemize}
For clarity, we first give a brief summary of the conclusions and provide the exact details at \ref{app experiments details}.

\subsection{Uncorrelated noise} 
\begin{figure*}[h]
\centering
     \includegraphics[width=0.33\textwidth]{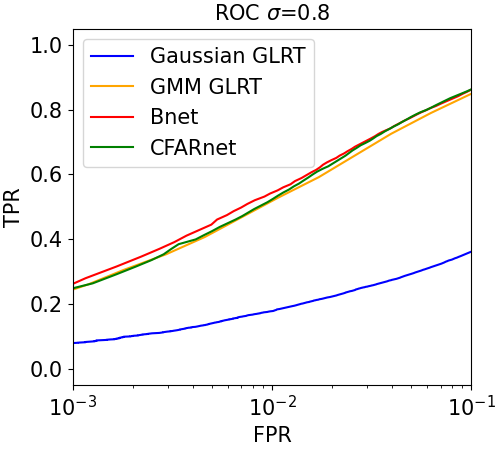}\includegraphics[width=0.33\textwidth]{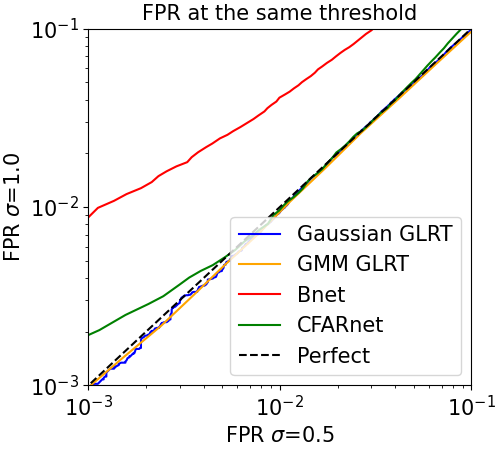}\includegraphics[width=0.33\textwidth]{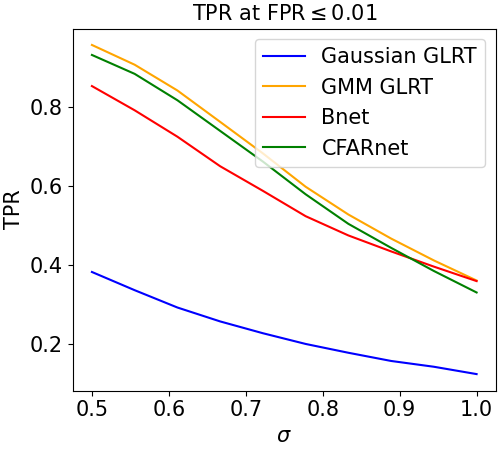} 
\\
\caption{Performance evaluation of the uncorrelated noise experiment (Right) The TPR as a function of the FPR for $\sigma=0.8$. (Middle) The FPR on $\sigma=1$ versus the FPR at $\sigma=0.5$ for different thresholds. (Left) The TPR as a function of $\sigma$ where the threshold is set to satisfy $\rm {FPR}\leq0.01$ for all values of $\sigma$. GMM GLRT, Bnet and CFARnet have similarly accuracy but Bnet is not CFAR. As a result, The worst case performance of Bnet are worse.
}
\label{fig:uncorrelated}%
\end{figure*}

In the first experiment we deal with problem in running example with non-Gaussian noise. As classical baselines we use the Gaussian GLRT and the exact non-Gaussian GLRT (denoted as GMM GLRT), Bnet and CFARnet. The setup is nearly asymptotic and GMM GLRT is the best in terms of accuracy and CFAR. Its main drawback is that it is computationally heavy due to the complicated likelihood. The Gaussian GLRT is much faster and CFAR but not very accurate. On the networks side, Bnet is very fast and leads to good ROCs, but is far from CFAR and performs bad under the worst case environment. Remarkably, CFARnet is nearly CFAR, nearly as accurate as Bnet and computationally efficient. These conclusions are summarized in Table \ref{table:non corrlated} and the results are shown in figure \ref{fig:uncorrelated}.

%Finally, we compare the TPR of the detectors when the threshold is set to fulfil FPR constraint to all values of $\sigma$ simultaneously. In such setup, the threshold is  needed to be set for the worst case FPR and thus the performance of non CFAR detectors are hardly injured. CFARnet and GMM-GLRT beats Bnet in this test.

\begin{table}[hbt!]
\vskip 0.15in
\begin{center}
% \begin{scriptsize}
\begin{sc}
\begin{tabular}{lcccc}
\toprule
Detector & ROC & CFAR & NP & SPEED \\
\midrule
Gaussian GLRT  & \xmark & \cmark &\xmark &\cmark \; (0.2 msec)  \\
GMM GLRT  & \cmark & \cmark &\cmark &\xmark \; (25 msec)\\
Bnet  & \cmark & \xmark &\xmark &\cmark \; (0.8 msec)\\
CFARnet   & \cmark & \cmark &\cmark &\cmark \; (0.8 msec) \\
\bottomrule
\end{tabular}
\end{sc}

% \end{scriptsize}
\end{center}
\label{table:non corrlated}
\caption{Detection in non-Gaussian noise: CFARnet approaches GMM GLRT performance with a much lower cost.}

\end{table}

\subsection{Locally correlated noise}
\begin{figure*}[h]
\centering
     \includegraphics[width=0.33\textwidth]{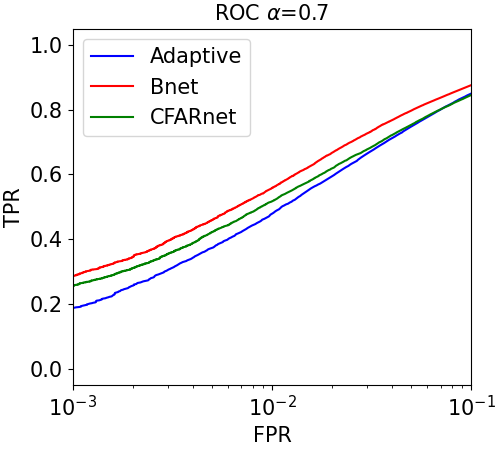}\includegraphics[width=0.33\textwidth]{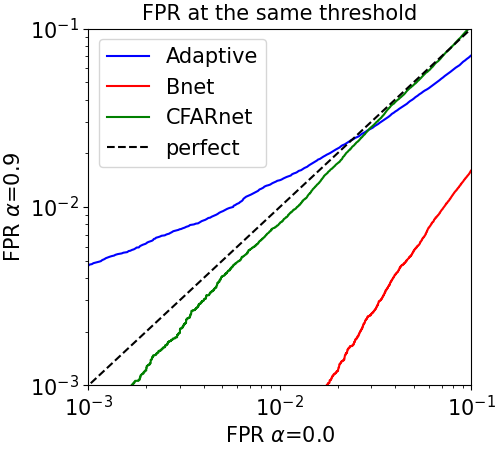}\includegraphics[width=0.33\textwidth]{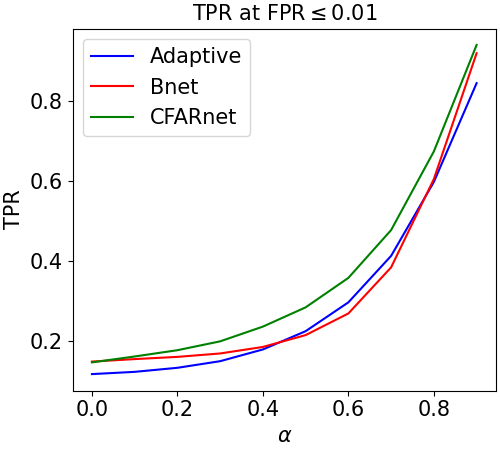} 

\caption{Performance evaluation of the locally correlated noise experiment.
(Left) the ROC curve for $\alpha=0.7$. CFARnet is worse then Bnet but better then Adaptive. (Middle) The FPR at the same threshold for $\alpha=0$ and $\alpha=0.9$. CFARnet is close to CFAR, while Bnet is far from CFAR. Adaptive is also less CFAR then CFARnet. (Left) TPR at FPR$\leq0.01$ constraint over all the regime of $\alpha$.  CFARnet dominates in the entire regime.}

\label{fig:locally corrlated}%
\end{figure*}

In our second experiment, the setting is as before but the noise is Gaussian yet correlated. 
Its covariance depends on a single unknown parameter $\alpha$:
\begin{align}
    \x = A \s + \w, ,\qquad \w \sim \mathcal{N}\(0,\Sig\),  
 \qquad \Sig_{ij}(\alpha) = \alpha^{|i-j|}.
\end{align}
In this setup, the GLRT is hard to compute, and therefore our ``Adaptive'' baseline uses a simple heuristic to estimate $\alpha$, and then plugs it into the known covariance GLRT. The results show that Bnet is very accurate but non-CFAR. CFARnet is slightly less accurate but CFAR and best in terms of NP. %is nearly CFAR and is more accurate then Adaptive, but slightly less accurate then Bnet. Finally, we compare the TPR of the detectors when the threshold is set to fulfil FPR constraint to all values of $\alpha$ simultaneously. CFARnet, beats both Bnet and Adaptive is this setup. 
These properties are summarized in table \ref{table:locally corrlated} and the results are shown in figure \ref{fig:locally corrlated}. 

\begin{table}[hbt!]

\vskip 0.15in
\begin{center}
% \begin{scriptsize}
\begin{sc}
\begin{tabular}{lccc}
\toprule
Detector & ROC & CFAR  & NP\\
\midrule
Adaptive & \xmark & \halfcheckmark & \xmark \\
Bnet  & \cmark & \xmark & \xmark \\
CFARnet   & \halfcheckmark & \cmark &\cmark \\
\bottomrule
\end{tabular}
\end{sc}
% \end{scriptsize}
\end{center}
\label{table:locally corrlated}

\caption{Detection in locally correlated noise: CFARnet has a slightly worse ROC but is best in terms of NP performance.}
\end{table}

\subsection{Secondary data}\label{exp corr}
\begin{figure*}[h]
\centering
\includegraphics[width=0.5\textwidth]{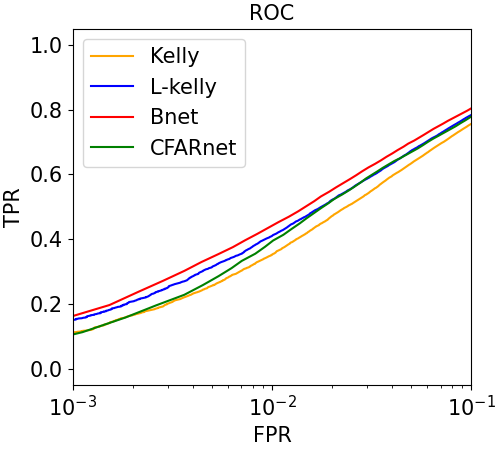}\includegraphics[width=0.5\textwidth]{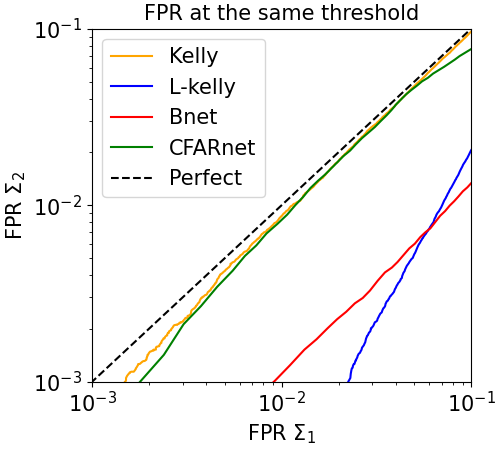}

\caption{Performance evaluation of the correlated noise with secondary input experiment. (Left) The ROC curves of different detection for some specific value of $\Sig$. Bnet better the CFARnet and L-Kelly, which are better then Kelly.(Right) FPRs under different covariance matrices. Kelly is CFAR and CFARnet is close to be CFAR. LAMF and Bnet are far from CFAR.}

\label{fig:corrlated}%
\end{figure*}

In our third experiment, we consider Gaussian noise with a completely unknown covariance $\Sig$. This is a classical problem in adaptive target detection \cite{kelly1986adaptive, robey1992cfar}. Following these works, we assume a secondary data of $n$ i.i.d. noise-only samples $\x_{aux}=(\w_1,...,\w_n)$. 
The classical baseline here is Kelly's detector which is known to be CFAR \cite{kelly1986adaptive}. The setting is non-asymptotic with dimension $5$ but only $n=20$ secondary samples. Therefore, we also compare to a regularized version of GLRT with diagonal loading denoted by L-Kelly.  It is more accurate but non CFAR. On the learning side, Bnet is even more accurate than L-Kelly but is also not CFAR. Finally, CFARnet is slightly less accurate than its competitors but is near-CFAR and performs best in the NP criterion.   %Finally, CFARnet is the best in terms of TPR where the threshold is set to satisfy an FPR constraint for  different parameters simultaneously.  
The properties are summarized in table \ref{table:corrlated} and the results are shown in figure \ref{fig:corrlated}.
\begin{table}[hbt!]

\vskip 0.15in
\begin{center}
% \begin{scriptsize}
\begin{sc}
\begin{tabular}{lccr}
\toprule
Detector & ROC & CFAR & NP  \\
\midrule
Kelly & \xmark & \cmark & \halfcheckmark  \\
L-Kelly & \halfcheckmark & \xmark & \xmark \\
Bnet  & \cmark & \xmark & \xmark \\
CFARnet   & \halfcheckmark & \cmark & \cmark \\
\bottomrule
\end{tabular}
\end{sc}
% \end{scriptsize}
\end{center}
\label{table:corrlated}

\caption{Detection with secondary data: CFARnet outperforms the competitors in terms of CFAR. In the appendix, we also report results that demonstrate that CFARnet is usually also best in terms of NP.}

\end{table}

\section{Discussion and future work}
In recent years, deep neural networks are become popular and are used for detection problems in many fields. While deep learning based detectors give remarkable improvements in accuracy, they are not CFAR and are thus unsuitable in many practical settings. 
We thus propose a method to train a deep learning detector that is CFAR, prove that it converges to GLRT in asymptotic settings, and show empirically that it results in CFAR detectors with a minor decrease in performance. 

It is important to address the limitations of the proposed framework. Fundamentally, our results demonstrate the asymptotic advantages of CFAR, but in finite samples there is an inherent tradeoff between fairness and accuracy that cannot be avoided. On the technical side, our implementation of CFARnet assumes that the synthetic dataset can be generated with full control of the parameters in the model. In addition, achieving CFAR at the low FPR regime is still challenging and requires large batch sizes. 
Future work can focus on relaxing the controllable dataset assumption, improve the efficiency of the methods in the low FPR regime. Together, the method can be evaluated in larger and more realistic settings.

\section*{Acknowledgment}
The authors would like to thank Yoav Wald for fruitful discussions and helpful insights. This research was partially supported by ISF grant number 2672/21.

\appendix 
\subsection{Details and full proof of Theorem 1}\label{proof of theorem 1}
% \anote{first, enumerate with A1-A99 all the necessary assumptions needed. second, enumerate all the 'with high probability' consequences. third, combine the assumptions+consequences with deterministic taylors to reach the required result (which will hold in high probability). clearly explain what high probability means}
The Proof is based on the Laplace's approximation \cite{wong2001asymptotic, tierney1986accurate, azevedo1994laplace, barndorff1989asymptotic} for the integral of functions with a "sharp peak" and express it in terms of the value of the function at it maximum. This allows us to express the relation between BLRT and GLRT. 

We begin by clearly specifying all the needed technical assumptions denoted by (Ax): 
\begin{enumerate}[label=(A\arabic*)]
    \item The input consists of $N$ i.i.d samples from $p(\x;\z)$.
    \item Weak signal: under the $y=1$ hypothesis, the true parameter $\z_{r_1}$ satisfies $\norm{\z_{r_1}-\z_{r_0}}=s/\sqrt{N}$ where $s>0$ is a constant.  
    \item The MLE converges to its asymptotic form \cite{kay1993fundamentals}:
    
\begin{align}\label{MLE asymptotic}
    \sqrt{N}\hat \z \sim \mathcal{N}\(\z, \I^{^{-1}}\), 
\end{align}

where $\boldsymbol{\mathcal{I}}$ is the Fisher Information Matrix (FIM) \footnote{Note the difference of our notation from \cite{kay1998fundamentals} where the joint $p(\x;\z)$ is used, compared to our definition where we use the FIM of a the single sample distribution $p(x;\z)$.} 
\begin{align} \label{FIM}
        \I = \EE{\frac{\partial^2\log\(p(x;\z)\)}{\partial \z^2}}=\(\begin{array}{cc}
        \I_{rr} &  \I_{rn}\\
        \I_{rn}^T & \I_{nn}
\end{array}\).
\end{align}
    \item The FIM $\I$ is not singular at the true parameter.
    \item The FIM is block diagonal, i.e., $\I_{rn}=\0$.
    \item The priors $p(\z_r)$ and $p(\z_n)$ are not zero at $\z_{r_0}$ and do not depend on $N$.
    \item The standard regularity conditions needed for Laplace approximation for marginal distribution as detailed in \cite{bilodeau2022stochastic} (see page 26-27 there). As explained there, these are standard asymptotic assumptions. Note also that some of them are related to the assumptions that we already stated, but we give them explicitly as we need them not only for the Laplace approximation. 
\end{enumerate}
Note that most of these assumptions are technical and standard, and follow \cite{kay1998fundamentals} and \cite{bilodeau2022stochastic}. The only non trivial assumption is (A5). It is satisfied in many classical models, e.g., in Gaussian distribution when the amplitude of the signal depend on the discriminative parameters and the covariance depends on the nuisance parameters. Future work may focus on relaxing this assumption.

% The exact dependence of $1/\sqrt N$ of the weak signal is not really needed, but we follow \cite{kay1998fundamentals}, which use this rate for constant GLRT performance with the increasing of $N$ \textcolor{blue}{I do not completely understand this, is the claim here like in the next sentence ("the assumption is not necessary but it simplifies the proof")? If yes, maybe we can group both sentences}. In addition, the i.i.d assumption can be replaced with a weaker assumption, but we use it for simplicity of the proof. 
The next step, is to use the above properties to approximate the GLRT and the BLRT, and to bound these approximations. Because the input $\x$ is probabilistic, the approximations can hold only in probability. Here we summarize the approximations that we use. As will detailed for each of them, the bounds hold with probability $p\rightarrow1$ when $N\rightarrow\infty$:
\begin{enumerate}[label=(B\arabic*)]
    \item The MLE converges to the true parameter:
    \begin{align}
        \hat \z &= \z + O\(\frac{1}{\sqrt{N}}\)%\frac{g_{\rm MLE }(\x)}{\sqrt{N}} 
        \label{determenstic bounds MLE} 
    \end{align}
    %where $g_{\rm MLE}(\x)$ is bounded. 
    Proof: This is immediate from the distribution of $\hat{\z}$ that is given in A3.
    % \item The distance between the MAP and MLE estimators satisfies:
    % \begin{align}
    %     \norm{\hat \z_r^{J} - \hat \z} = O\(\frac{1}{{N}}\)
    %     \label{determenstic bounds MAP},
    % \end{align}
    \item The log likelihood at the MAP and the log likelihood at the MLE are related by:
         \begin{align}
        \sumn p(x_i;\hat \z_r^{J}) = \sumn p(x_i;\hat \z) + O\(\frac{1}{{N}}\)
        \label{determenstic bounds MAP},
    \end{align}
    Proof: See \ref{proofB2}.
    \item The marginal distribution can be computed using Laplace's approximation:
\begin{align}
    p_1(\x) &= \tilde p_1(\x)\(1 +  O\(\frac{1}{{N}}\)\)% \frac{g_{\rm LA }(\x)}{N}\) 
    \label{determenstic bounds Laplace}.
\end{align}
    where $p_1(\x)$ is the exact marginal density:
        \begin{align}
    p_1(\x) = \int \prodn p(x_i,\z)p(\z)d\z,
\end{align}
and $\tilde p_1(\x)$ is the Laplace approximated density:
\begin{align}\label{laplace marginal}
     \tilde p_1(\x) = \prodn p(x_i;\hat \z^{J})p(\hat \z^{J})\sqrt{\frac{\(2\pi\)^{d_r}}{N^{d_r} \Det\(\tilde \I(\hat \z^{J})\)}},
 \end{align} 
 where
 \begin{align}
     \tilde \I = \frac{1}{N}\sum_{i=1}^N\frac{\partial^2\log(p(x_i,\z))}{\partial \z^2}
 \end{align}
 is the Hessian.

Proof: Using the assumptions in A7, we use the results of \cite{bilodeau2022stochastic, bilodeau2022tightness}. 
\item The Hessian converges to the FIM: $\tilde \I \rightarrow \I$ and thus for large enough $N$, the Hessian is not singular (by A4).
Proof: The hessian is given by the sample average of IID variables, thus  by the weak law of large numbers, it converges in probability to their mean.
\end{enumerate}    
From now on, we use these bounds without explicitly mention that they hold only in probability.
We turn to proving Theorem 1. We start with the case of no nuisance parameters. GLRT (\ref{GLRT}) and BLRT (\ref{BLRT}) can be written as
\begin{align}
T_{\rm GLRT}(\x) &=2\sum_{i=1}^N\log\(p(x_i;\hat \z_r)\)-2\sumn\log\(p(x_i; \z_{r_0})\) \label{GLRT appendix}\\
T_{\rm BLRT}(\x) &=2\log\int \prodn p(x_i;\z_r)p(\z_r)d\z_r \nonumber \\ 
 &- 2\sumn \log\(p(x_i; \z_{r_0})\) \label{BLRT appendix}
\end{align}
Taking the log of the marginal after Laplace approximation (\ref{determenstic bounds Laplace}) gives:
\begin{align}\label{log marginal}
    \log\(p_1(\x)\) &= \log\int \prodn p(x_i;\z_r)p(\z_r)d\z_r \nonumber \\ 
    & = \sumn\log\(p(x_i;\hat \z_r^{J})\) + \log\(p(\hat \z_r^{J})\) \nonumber \\
    &+\frac{1}{2}d_r\log\(\frac{2\pi}{N}\) - \frac{1}{2}\log\( \Det\(\tilde \I(\hat \z_r)\)\) \nonumber \\
    &+ \log\(1+O\(\frac{1}{N}\)\) 
\end{align}
%the Taylor theorem states that for any differentiable function $f(\z)$, 
% \begin{align}\label{taylor}
%     \log f(\z_2) = \log \(f(\z_1)\) + \frac{\nabla f(\z')}{f(\z')}\cdot\(\z_2 -\z_1)
% \end{align}
 %where $\z'$ is some point between $\z_1$ and $\z_2$. Thus, if $f(\z)$ is not zero in the region between $\z_1$ and $\z_2$, we can say that:
%  Using, B2, we get
%  In our next step, we approximate the log likelihood at the MAP with the log likelihood at the MLE. To get a bound on the error of the approximation, we use Taylor theorem.
%  By Taylor, for any close enough $\z_1$ and $\z_2$, and for any function three-times differentiable function $q$ we have that:
%  \begin{align}\label{taylor O}
%     q(\z_2) = q(\z_1) + \nabla q(\z_1)\cdot(\z_2-\z_1) + O\(\norm{\z_2-\z_1}^2\) 
% \end{align}
% for any $q$ as long as q=O(1). Now note that the log likelihood is $O(N)$, and thus we will use $q(\z)=\frac{1}{N}\sumn \log p(x_i;\hat \z)$. since $\hat{\z}$ is the maximum of $q$, we have that $\nabla q(\hat{\z})=0$ and thus
% \begin{align}\label{map to MLE 0}
%     \frac{1}{N}\sumn \log \(p(x_i;\hat \z^J)\) = \frac{1}{N}\sumn \log \(p(x_i;\hat \z)\) + O\(\norm{\hat \z - \hat \z^J}^2\)
% \end{align}
% Plugging (B2), and multiplying (\ref{map to MLE 0}) by $N$ gives:
%  \begin{align}\label{map to MLE}
%      \sumn \log \(p(x_i;\hat \z^{J})\) = \sumn \log \(p(x_i;\hat \z)\) + O\(\frac{1}{N}\).
%  \end{align}
%  In addition, we use:
% \begin{align}\label{log 1+a}
%     \log\(1+O\(\frac{1}{N}\)\) = O\(\frac{1}{N}\),
% \end{align}
% Plugging (\ref{map to MLE}), (\ref{GLRT appendix}) and (\ref{log 1+a}) into 
 Now we use (B2) and plug (\ref{determenstic bounds MAP}) into (\ref{log marginal}) and then to \ref{BLRT appendix}. In addition we use:
\begin{align}\label{log 1+a}
    \log\(1+O\(\frac{1}{N}\)\) = O\(\frac{1}{N}\),
\end{align}
and get:
 \begin{align} \label{BLRT with MLE}
     T_{\rm BLRT} &= 2\sum_{i=1}^N\log\(p(x_i;\hat \z_r)\)-2\sumn\log\(p(x_i; \z_{r_0})\) \nonumber \\
     &+d_r\log\(\frac{2\pi}{N}\) + 2\log\(p(\hat \z_r^{J})\)\nonumber \\
     &- \log\( \Det\(\tilde \I(\hat \z_r)\)\) + O\(\frac{1}{N}\).
\end{align}
The first two terms are exactly the GLRT.

Next, we use Taylor theorem to approximate the terms that depend on $\hat \z_r$ and $\hat \z_r^J$ by their values at $\z_0$. For any close enough $\z_1$ and $\z_2$, and for any two-times differentiable function $q$ we have that:
 \begin{align}\label{taylor O}
    q(\z_2) = q(\z_1) + O\(\norm{\z_2-\z_1}\) 
\end{align}
Using (A2) and (B1)-(B2) yield
\begin{align}\label{small signal}
    \hat \z_r &=  \z_{r_0} + O\(\frac{1}{\sqrt N}\) \nonumber \\
    \hat \z_r^J &=  \z_{r_0} + O\(\frac{1}{\sqrt N}\).
\end{align}
Plugging (\ref{small signal}) into (\ref{taylor O}) together with (A4) and (A6) gives:
\begin{align}
     \log\( \Det\(\tilde \I(\hat  \z_{r})\)\) &=   \log\( \Det\(\tilde \I( \z_{r_0})\)\) + O\(\frac{1}{\sqrt N}\) \nonumber \\
    2\log\(p(\hat  \z_{r}^J)\) &= 2\log\(p( \z_{r_0})\) + O\(\frac{1}{\sqrt N}\)
\end{align}

under both $y=0$ and $y=1$. 
% \begin{align}דדדדדדדדדדדדדדדדדדדדדדדדדדדדדדדד
%     \log\(1+\frac{g_{\rm LA}}{N}\) = O\(\frac{1}{N}\),
% \end{align}
Plugging these into (\ref{BLRT with MLE}) gives: 

 \begin{align}
     T_{\rm BLRT} &= T_{\rm GLRT} +d_r\log\(\frac{2\pi}{N}\) + 2\log\(p( \z_{r_0})\)\nonumber \\
     &- \log\( \Det\(\tilde \I( \z_{r_0})\)\) + O\(\frac{1}{\sqrt N}\).
 \end{align}

Finally, because $\z_{r_0}$ is known, we get that

 \begin{align}
     T_{\rm BLRT} &= T_{\rm GLRT} + const. + O\(\frac{1}{\sqrt N}\)
 \end{align}

The case with nuisance parameters is very similar.
Here the GLRT is given by:
\begin{align}
    T_{\rm GLRT} = \frac{1}{2}\frac{\log\prod_{i=1}^N p(x_i,\hat \z)}{\log\prod_{i=1}^Np(x_i,\z_{r_0}, \hat \z_{n_0})} 
\end{align}
where $\hat \z= (\hat \z_r, \hat \z_n)$ is the MLE and $\hat \z_{n_0}$ is the constrained MLE where $\z_r=\z_{r_0}$ and is given by \cite{kay1998fundamentals}:
\begin{align}
    \hat \z_{n_0} \approx \hat \z_{n} + \I_{nn}^{-1}\(\hat \z\)\I_{nr}\(\hat \z\)\(\hat \z_r-\z_{r_0}\).
\end{align}
Thus, using B1, the unrestricted MLE of $\hat \z_n$ and the restricted MLE $\hat \z_{n_0}$ satisfy in high probability:
\begin{align}
    \hat \z_n =\z_n +O\(\frac{1}{\sqrt N}\) \nonumber \\
    \hat \z_{n_0} =\z_n +O\(\frac{1}{\sqrt N}\),
\end{align}
under both $y=0$ and $y=1$. 
The BLRT is given by:
\begin{align}
    T_{\rm BLRT}(\x) =\frac{1}{2}\log\(\frac{\tilde p_1(\x)}{\tilde p_0(\x)}\) 
\end{align}
where 
\begin{align}\label{int_nuisance}
    \tilde p_1(\x) &= \int \prodn p(x_i;\z_r, \z_n) p( \z_r) p(\z_n)d\z_r d\z_n \nonumber \\
    \tilde p_0(\x) &= \int \prodn p(x_i;\z_{r_0}, \z_n)p( \z_n) d\z_n.
\end{align}
Using the Laplace approximation (\ref{determenstic bounds Laplace}) for both $\tilde p_1(\x)$ and $\tilde p_0(\x)$, and Taylor expansions, the BLRT can be expressed as:
\begin{align}
    T_{\rm BLRT}(\x) = T_{\rm BLRT}(\x) + \rho(\z_{r_0},\z_n) + O\(\frac{1}{\sqrt N}\) + const.
\end{align}
where 
\begin{align}
    \rho(\z_{r_0},\z_n) = \log\(\frac{\Det\(\I(\z_{r_0},\z_n)\)}{\Det\(\I_{nn}(\z_{r_0},\z_n)\)}\).
\end{align}
Here $\z_n$ is the unknown true value of the nuisance parameter vector. 

\subsection{Proof of theorem 2}
 GLRT is asymptotically CFAR and if the FIM is block diagonal its performance with unknown nuisance parameters is the same as if they where known (and just $\z_r$ is unknown) (see \ref{appBlockDiag}).
Thus, GLRT has the best expected (over $\z_r$) TPR for any given FPR, for any $\z_n$. 
Now, we write the probability of error as the sum of the expected FPR and the expected false negative rate
(FNR, which is 1-TPR): 
Given a test $(T(\x),\gamma)$, we denote the expected FPR for a given $\z_n$ by:
\begin{equation}
        \alpha(T, \gamma,\z_n)={\rm Pr}\left(T(\x)>\gamma|y=0,\z_n\right),
\end{equation}
and similarly the expected FNR for a given $\z_n$ by:
\begin{equation}
     \beta(T, \gamma, \z_n)={\rm Pr}\left(T(\x)<\gamma|y=1,\z_n\right).
\end{equation}
The objective of CLRT is therefore:
\begin{equation}
        l_{0-1}(T, \gamma) =p_0\EE{\alpha(T, \gamma, \z_n)} +p_1\EE{\beta(T, \gamma, \z_n)},
\end{equation}
where the expectations are over $p(\z_n)$.

We now show that the GLRT with some threshold $\gamma$, is a solution to (\ref{CFAR optimization}) which can be written as:
\begin{align}
    {\rm min}_{T(\x),\gamma} \quad & l_{0-1}(T,\gamma) \nonumber \\
    {\rm{s.t.}} \quad & \alpha(T, \tilde \gamma,\z_n)=\alpha(T, \tilde \gamma,\z_n') \quad \forall {\z_n, \z_n',\tilde \gamma},  
\end{align}
Note that the constraint must be satisfied for all $\tilde \gamma$ and not just for the optimal $\gamma$.

Due to the CFAR constraint, the FPR is constant with respect to  $\z_n$ and so is its expectation:
\begin{equation}
    \EE{\alpha(T,\gamma, \z_n)}=\alpha(T,\gamma),
\end{equation}
where $\alpha(T,\gamma)$ is the FPR of the test on any value of  $\z_n$. Thus the Bayesian 0-1 loss can be written as:
\begin{equation}\label{Bayesian 0-1 loss}
    l_{0-1}(T, \gamma) =p_0\alpha(T,\gamma) + p_1\EE{\beta(T,\gamma,\z_n))}.
\end{equation}
The best threshold for any detector can be therefore found by minimizing (\ref{Bayesian 0-1 loss}) with respect to $\gamma$.
Specifically we denote the optimal threshold of the GLRT detector by $\gamma^*$, and denote the corresponding FPR  as $\alpha^*$.

Now we prove that any other CFAR detector gives a larger or equal Bayesian 0-1 loss.
We assume that there exist a detector $(T'(\x,\z_n),\gamma')$ that has FPR of $\alpha'$ for any value of $\z_n$. Its Bayesian loss is given by: 
\begin{align}
   l_{0-1}(T', \gamma') &= p_0\alpha'+p_1\EE{\beta(T', \gamma', \z_n)} \nonumber\\
            & \geq p_0\alpha'+p_1\EE{\beta(T_{\rm GLRT}, \gamma'_{\rm GLRT}, \z_n)} \nonumber \\
            &=l_{0-1}(T_{\rm GLRT}, \gamma'_{\rm GLRT}) \nonumber \\
            &\geq l_{0-1}(T_{\rm GLRT}, \gamma^*),
\end{align}
where $\gamma'_{\rm GLRT}$ is the threshold that gives FPR of $\alpha'$ to the GLRT. The first inequality is due to the optimality of GLRT among detectors that have FPR $\alpha'$ for any value of $\z_n$. The second inequality is due to the optimality of the threshold $\gamma^*$ for the GLRT detector.

In conclusion, the GLRT with threshold $\gamma^*$ gives the minimum Bayesian loss among all the detectors that have constant false alarm rate over $\z_n$. In other words, GLRT and CLRT are equivalent, completing the proof.

\subsection{Proof of B2}\label{proofB2}
First we state the following lemma:
\begin{lemma}
Let $f$ and $h$  be thrice differentiable functions with unique maxima such that $|h(\z)-f(\z)|=O(1/N)$ for all $\z$. Assume also that the Hessian of $f$ at its maximum is $O(1)$  and is bounded away from zero. Denote $\hat \z^f=\arg\max f(\z)$, $\hat \z^h=\arg\max h(\z)$, then: 
\begin{align}
    f(\hat \z^h) = f(\hat \z^f) + O\(\frac{1}{N^2}\).
\end{align}
\end{lemma}
\begin{proof}
For simplicity, we only consider the scalar case. Define $g(\z)=h(\z)-f(\z)$. Due to Taylor, since $\hat \z^f$ is a local maximum of $f$, the derivative of $f$ zero at $\hat \z^f$ and for any point $\z$ near $\hat \z^f$:
\begin{align}\label{taylor f}
    f(\z) = f(\hat \z^f) + \frac{1}{2}f''(\hat \z^f)\(\Delta\z\)^2 + O\((\Delta\z)^3\) 
\end{align}
where $\Delta\z=\z -\hat \z^f$. On the other hand, the derivative of $g$ does not necessarily vanish 
\begin{align}
    g(\z) = g(\hat \z^f) + g'(\hat \z^f) (\Delta\z) + \frac{1}{2}g''(\hat \z^f)(\Delta\z)^2 + O\((\Delta\z)^3\).
\end{align}
Thus, near $\hat \z^f$, we have
\begin{align}
   h(\z) =& \left[\frac{1}{2}f''(\hat\z^f)+\frac{1}{2}g''(\hat\z^f)\right](\Delta\z)^2 + g'(\hat\z^f)(\Delta\z)\nonumber\\
   &+ [f(\hat\z^f)+g(\hat\z^f)]  + O\((\Delta\z)^3\)
\end{align}
and the maximum of $h$ satisfies,
\begin{align}
     h'(\z) = \left[f''(\hat\z)+g''(\hat\z)\right](\Delta\z) +  g'(\hat\z) + O((\Delta\z^2)) = 0.
\end{align}
Isolating $\Delta\z$ gives:
\begin{align}\label{delta z}
    \Delta\z=\frac{g'(\hat\z)+O((\Delta\z^2))}{f''(\hat\z)+g''(\hat\z)}.
\end{align}
Now, $f''$ is $O(1)$ and $g'$ and $g''$ are $O(1/N)$, Thus:
\begin{equation}\label{delta z}
    \Delta\z = O\(\frac{1}{N}\).
\end{equation}
Note that although the expression for $\Delta x$ is in implicit because of the $O((\Delta\z^2))$ term, this term does not influence the result as if $\Delta\z$ is not $O(1/N)$, the equality can not hold.
Plugging (\ref{delta z}) into (\ref{taylor f}) gives:
\begin{align}
     f(\hat \z^h) = f(\hat \z^f) + O\(\frac{1}{N^2}\)
\end{align}
\end{proof}
 % \anote{this paper https://ieeexplore.ieee.org/stamp/stamp.jsp?tp=&arnumber=1306646 has the exact asymptotic analysis of PML and all the necessary assumptions for it}.
% \anote{this last result could be a nice lemma stating the non-obvious 1/N and 1/sqrtN relation between the true, ml and map}
Now we apply the lemma on
\begin{align}
    f(\z)&=\frac{1}{N}\sumn\log\(p(x_i;\z)\) \nonumber\\
    h(\z)&=f(\z) + \frac{1}{N}\log \(p(\z)\).
\end{align}
The log of the prior $\log (p(\z))$ is $O(1)$ and thus $f$ and $h$ satisfy $|h(\z)-f(\z)|=O(1/N)$.
The second derivative of $f$ converges in high probability to the FIM (B4) which is non zero and is $O(1)$, Thus $f$ and $g$ satisfy the conditions in high probability. Thus, $\hat \z$ is the maximum of $f(\z)$, $\hat \z^{J}$ is the maximum of $h(\z)$ and:
\begin{align}
    \frac{1}{N}\sumn p(x_i,\hat \z^J) = \frac{1}{N}\sumn p(x_i,\hat \z) + O\(\frac{1}{N^2}\).
\end{align}
Multiplying all by $N$ gives (\ref{determenstic bounds MAP}), completing the proof of B2.

\subsection{Performance of GLRT when the FIM is block-diagonal}\label{appBlockDiag}
In the proof of theorem 2, we used the property that, when the FIM is block diagonal, the performance of GLRT with unknown nuisance parameters is the same as if they were known. Here we show this property explicitly using the asymptotic distribution of GLRT.
The asymptotic distribution of GLRT is \cite[p. 206]{kay1998fundamentals}:
\begin{align}
    T_{\rm GLRT} \sim \begin{cases}
        \chi_{d_r}^2, & y=0 \\
        \chi_{d_r}^2(\lambda), & y=1
    \end{cases}    
\end{align}
where $\chi_{d_r}^2,$ is the chi-squared distribution and $\chi_{d_r}^2(\lambda)$ is the non-central chi-squared distribution. If the there are no nuisance parameters, the parameter $\lambda$ is:
\begin{equation}\label{asy} 
    \lambda_{\text{known}} = N \(\z_r-\z_{r_0}\)^T \I(\z_{r_0},\z_n)\(\z_r-\z_{r_0}\).    
\end{equation}
where $\z_n$ are the known parameters of the distribution (usually the FIM is not written as a function the known parameters, but here we write this dependence explicitly to compare it to the unkown case).

If there are unknown nuisance parameters, the parameter is: 
\begin{align}\label{asy_non} 
    \lambda_{\text{unknown}} &= N (\z_r-\z_{r_0})^T (\I_{rr}(\z_{r_0},\z_n)\nonumber \\
    &-\I_{rn}(\z_{r_0},\z_n)\I^{-1}_{nn}(\z_{r_0},\z_n)\I_{rn}^T(\z_{r_0},\z_n))(\z_r-\z_{r_0}).
\end{align}
Here $\z_n$ are the true but unknown nuisance parameters.
Under $y=0$, the distributions of GLRT are the same in both cases. Under $y=1$, if $\I_{rs}=0$, (\ref{asy_non}) reduces to (\ref{asy}) and $\lambda_{\text{known}}=\lambda_{\text{unknown}}$. Thus the detection performance is the same in both cases.

\subsection{Statistical distances and MMD}\label{app_dist}

Maximum Mean Discrepancy (MMD) is a statistical distance defined as \cite{gretton2012kernel}:
\begin{align}
        d_{\rm{MMD}}(X;Y) = \E[k(X,X')]  + \E[k(Y,Y')] - 2 \E[k(X,Y)]
\end{align}\label{mmd_def}
where $X$ and $X'$ are independent and identically distributed (i.i.d.), and so are $Y$ and $Y'$. The function $k(\cdot,\cdot)$ is a characteristic kernel over a reproducing kernel Hilbert space, e.g.,  the Gaussian Radial Basis Function (RBF).
Recent advances in deep generative models allow us to optimize distances as MMD in an empirical and differentiable manner \cite{li2015generative}. For this purpose, we need to represent each distribution using samples drawn from it. Let $\{X_i\}_{i=1}^N$ and $\{Y_i\}_{i=1}^N$ be i.i.d. realizations of $X$ and $Y$, respectively. Then, an empirical version of the MMD can be used where the expectations in (\ref{mmd_def}) are replaced by their empirical estimates.
More advanced metrics can be obtained using  
% \begin{align}
%     \hat R_{\rm{adv}}(\hat{T})= \sum_{i=1}^N\max_{g(\cdot)}\left|\sum_{j=1}^Mg(T_{ij})-\sum_{j=1}^Mg(\tilde{T}_{ij})\right|^2
% \end{align}
the tools of generative adversarial networks (GANs). In this paper, we only deal with distances between scalar random variables and simple MMD distances suffice.  


\subsection{Experiments details}\label{app experiments details}
 
In each experiment, Bnet and CFARnet have identical architecture and both of them are trained using stochastic optimization \cite{shalev2014understanding}, until the loss function reaches a plateau. The classification loss is cross-entropy.
CFARnet is implemented by generating two different batches in each step, one for the classification loss and another one for the CFAR penalty. The batch of the classification loss includes i.i.d. samples with random $y$ and $\z$ according to the chosen prior and while the batch of the CFAR penalty includes only samples of $y=0$ from two different values of $\z$ that are sampled i.i.d from the prior. MMD is used for the CFAR penalty.
An important practical issue is that the MMD loss is governed by the high FPR regime which is usually not interesting. To avoid this, in each step we generate 20 times larger batch and then take only 5\% of the samples with the highest score to the MMD loss. This is reminiscent of the hard negative sampling methods that are common ni visual object detection \cite{liu2016ssd}.  Another important issue is that the batch size form the MMD loss should be large enough so that the samples represent the distribution well enough. We found that batch size of 500 (that is 10000 before taking 5\%) was good enough.
The choice of the value of $\lambda$ was done by a standard hyperparamter search. In addition, training CFARnet with a positive $\lambda$ from scratch leads often to unstable learning, and a good practice that we found is to set $\lambda=0$ in the first iterations and only then increase it. Finally, we
used the PyTorch library \cite{NEURIPS2019_9015} and  Adam \cite{kingma2017adam} for the optimization.  
% \paragraph{Performance Evaluation}
% The goal of the performance evaluation is to measure both the accuracy and the degree of CFAR of the methods. The accuracy is presented by the ROC curve in different values of the unknown parameters. The degree of CFAR is visualized by the FPR at one value of parameter versus the FPR at a different value of the parameter, on different thresholds. 
% Finally, we measure the TPR for an FPR constraint of less then 0.01 for all values of the unknown parameters. This measure demonstrate the realistic cases when the threshold is chosen such that the FPR constraint is satisfied for all values of the unknown parameters. In all of the experiments, the parameters in test set lie in the region that was used for training.

\subsubsection{Non-Gaussian noise experiment}\label{exp uncor} 
The first experiment is based on running example of unknown target amplitude and unknown noise scaling as defined in (\ref{DC})-(\ref{params}) where the noise contains outliers. The noise is modeled by the following non-Gaussian distribution:
\begin{align}
    p(n_i)= (1-\epsilon)\mathcal{N}(0,1) + \epsilon \mathcal{N}(0,100).
\end{align}
The number of samples is 40 and the valid region of the unknown parameters is:
\begin{align}\label{params_traininig}
    -1\leq A\leq 1,\qquad 0.5\leq \sigma\leq 1.
\end{align}
The test set contains 100,000 samples for each 10 linearly spaced values of $\sigma$ from 0.5 to 1 and the amplitude of the signal is $A=0.5$.

The first baseline is the Gaussian GLRT (that is, assuming Gaussian noise). It has a simple closed form solution (\ref{GLRT uncorelated}) and is known to be CFAR.
 The second method is GMM GLRT, which is the GLRT associated with the true noise distribution, where the optimizations are performed using the common Expectation-Maximization (EM) algorithm \cite{dempster1977maximum}. In order to achieve efficient execution on a GPU, the implementation has been vectorized, thus the EM algorithm executes for a predetermined number of steps, (with 5 steps being determined as the optimal balance between computational speed and accuracy).

The architecture of CFARnet and Bnet is based first on element-wise functions, that are implemented by a convolution layer with kernel size of 1 \cite{lin2013network}. After two such layers (with 50 output channels and a ReLU non linearity after each of them), an average is preformed over all the elements, giving a vector of size 50. These are followed by a fully connected layer of size 50 with ReLU non linearities and a final linear classification layer. The training batch size is 500 and the CFAR loss parameter in CFARnet is $\lambda=0.1$. The prior distributions for generating the training data are $N(0,1)$ for $A$ and a uniform distribution in $[0.5,1]$ for $\sigma$.

\subsubsection{Locally correlated noise experiment}
Here we consider a known signal (with unknown amplitude $A$) in a correlated Gaussian noise.
The covariance of the noise depend on a single unknown parameter $\alpha\in[0,0.9]$:
\begin{align}
    \x = A \s + \w, ,\qquad \w \sim \mathcal{N}\(0,\Sig\),  
 \qquad \Sig_{ij}(\alpha) = \alpha^{|i-j|}.
\end{align}
% The goal is to decide whether $A$ is equal to zero or not.
% \paragraph{Classical baselines} 
The classical  ``Adaptive'' baseline is based on the GLRT of known covariance $\Sig$ \cite{robey1992cfar}
\begin{equation}\label{glrt_known_covariance}
    T(\x) = \frac{\left(\s^T \Sig^{-1}\x\right)^2}{\s^T \Sig^{-1}\s}.
\end{equation}
where the parameter $\alpha$ is estimated heuristically. We use the following steps:
\begin{itemize}
    \item Estimate $A$ by ML assuming $\alpha=0$.
    \item Define $\z = \x - A\s$.
    \item Estimate $\alpha$ by method of moments: $\hat \alpha=\frac{n}{n-1}\frac{\sum_{i=1}^{n-1}\z_i\z_{i+1}}{\sum_{i=1}^{n}\z_i^2}$
    \item Plug $\Sig(\hat \alpha)$ into (\ref{glrt_known_covariance}).
\end{itemize}
The test set contains 100,000 samples for each 10 linearly spaced values of $\sigma$ from 0 to 0.9. The amplitude of the signal is $A=0.4$.
%\paragraph{Architecture and training parameters}

The architecture for CFARnet and Bnet is a convolution neural network \cite{krizhevsky2012imagenet} based on 3 convolution layers with a ReLU activation with 20 channels (kernal sizes of [3,2,2]), a hidden fully connected later of size 400 with a RelU activation and a linear classification layer. The prior distributions for generating the training data are $N(0,1)$ for $A$ a uniform distribution in $[0,0.9]$ for $\alpha$.
The training batch size is 100 and the CFAR loss parameter in CFARnet is $\lambda=1$. 

% \paragraph{Results}
% The results are summarized in Fig. \ref{fig:cor}.
% The right plot shows the ROC curve for $\alpha=0.7$. CFARnet is worse then Bnet but better then Adaptive. The middle plot shows the FPR at the same threshold. CFARnet is close to CFAR, while Bnet is far from CFAR. Adaptive is also less CFAR then CFARnet. Finally, (left plot) CFARnet dominates the TPR at FPR$\leq0.01$ constraint over all the regime of $\alpha$.

\subsubsection{Secondary data experiment}\label{exp corr}
%\paragraph{Problem formulation}
Here we consider the detection of a known signal $\s\in \mathbb{R}^d$ with unknown amplitude in Gaussian noise with unknown covariance $\Sig$. This is a classical problem in adaptive target detection \cite{kelly1986adaptive, robey1992cfar}. Following these works, we assume a secondary data of $n$ i.i.d. noise-only samples $\x_{aux}=(\w_1,...,\w_n)$. Together the observations can be modelled as
\begin{align}
    & \x = A\s + \w_0\nonumber\\
    &\x_i = \w_i \quad i=1,\cdots, n
\end{align}
where
\begin{equation}
        \w_0,\w_i\sim {\mathcal{N}}({\mathbf{0}}, \Sig)
\end{equation}
In terms of the standard notations of detection from earlier, The vector $\z$ includes both $A$ and all the elements in the matrix $\Sig$. The goal is to decide between
\begin{align}
    &y=0:\quad A=0\nonumber \\
    &y=1:\quad A\neq 0.
\end{align}
We set $d=5$ and $n=20$. The test set contains 200,000 samples from  different covariance matrices that was sampled from the Wishart distribution ${\mathcal{W}}_{2d}\(\boldsymbol{I}_{d\times d}, d\)$. For visiblity, we report the results on two samples of covariance matrices, but the results where examined on about 50 samples, all had simmiliar behaviour to thre reported samples. The amplitude of the signal is $A=1$.

Kelly's detector is defined as \cite{kelly1986adaptive}:
    \begin{equation}
    T_{\rm Kelly}(\x) = \frac{\left(\s^T \hat \Sig^{-1}\x\right)^2}{(\s^T\hat \Sig^{-1}\s)(1+\frac{1}{n}\x^T\hat \Sig^{-1}\x)},
\end{equation}
where 
\begin{align}\label{sample covariance}
    \hat\Sig = \frac 1n \sum_{i=1}^n \w_i\w_i^T,
\end{align}
is the sample covariance of the secondary data. We also experimented with the famous adaptive matched filter (AMF) detector \cite{robey1992cfar} which performed similarly. Both detectors are CFAR but sub-optimal when $n$ is small. In such settings, it is common to use L-Kelly which plugs in regularized covariance estimators \cite{ledoit2004well,abramovich2007modified}: 
\begin{equation}
    \hat \Sig_{\lambda} = \hat \Sig +\lambda\boldsymbol{I}.
\end{equation}
where $\lambda$ is a diagonal loading hyperparameter ($\lambda=3$ is our experiments).

The architecture for CFARnet and Bnet is a based on non-linear features the are the ingredients of the classical Kelly and (L)AMF detectors:
\begin{align}
    f^\lambda_1(\x) &=\s^T\hat \Sig_{\lambda}^{-1}\x \nonumber \\
    f^\lambda_2(\x) &=\s^T\hat \Sig_{\lambda}^{-1}\s \nonumber \\
    f^\lambda_3(\x) &=\x^T\hat \Sig_{\lambda}^{-1}\x.
\end{align}
We use 10 different linearly spaced values of $\lambda$ between 0 and 0.3. All the 30 features are concatenated to a single vector and are fed into a fully connected neural network with a single hidden layer of size 100 and a ReLU non linearity. The prior $A$ is set to be $\mathcal{N}(0,1)$ and the prior for $\Sig$ is a wishard distribution ${\mathcal{W}}_{2d}$. The batch size is 100 and the parameter $\lambda$ is set to be 0.2. 

The main results are summarized in Table \ref{table:corrlated} and figure \ref{fig:corrlated} within the text. We also performed experiments with respect to the NP metric which analyzes the TPR for an FPR constraint and different nuisance parameters. The results obviously depend on the parameters and change across the simulations. To give a taste of the typical behaviour, Table \ref{table:corrlated wc} reports the TPR for five randomly chosen nuisance covariances. There are five realizations from the same distribution of the training set. In four out of five experiments, CFARnet outperforms its competitors. 

% \paragraph{Results}
% The results are summarized in Fig. \ref{fig:aux}.
% In the left, plot  ROC curves of different detection for some specific value of $\Sig$. The order form the best to the least is Bnet, L-Kelly, CFARnet, Kelly.
% The right plot shows the FPRs under different covariance matrices. Kelly is CFAR and CFARnet is close to be CFAR. LAMF and Bnet are far from CFAR.
% Finally, in table \ref{table:corrlated wc} we show TPR where the threshold is set to satisfy the an FPR constraint of 0.01 on 20 randomly sampled covariance matrices (). The TPR is averaged over the 20 covariance matrices.  

% \begin{table}[t]
% \caption{Average TPR for ${\rm FPR}\leq0.01$ }

% \vskip 0.15in
% \begin{center}
% % \begin{scriptsize}
% \begin{sc}
% \begin{tabular}{lccr}
% \toprule
% Detector & Average TPR \\
% \midrule
% Kelly & 0.62    \\
% L-Kelly & 0.55\\
% Bnet  & 0.53\\
% CFARnet   &  \textbf{0.64}\\
% \bottomrule
% \end{tabular}
% \end{sc}
% % \end{scriptsize}
% \end{center}
% \vskip -0.1in
% \label{table:corrlated wc}

% \end{table}

\begin{table}[t]
\caption{TPR for ${\rm FPR}\leq0.01$ }

\vskip 0.15in
\begin{center}
% \begin{scriptsize}
\begin{sc}
\begin{tabular}{lccccr}
\toprule
Detector & TPR 1 & TPR 2  & TPR 3 & TPR 4 & TPR 5 \\
\midrule
Kelly   & 0.35          & 0.84          & 0.84          & 0.11          & 0.64          \\
L-Kelly & 0.36          & 0.75          & 0.56          & \textbf{0.13} & 0.47          \\
Bnet    & 0.3           & 0.69          & 0.65          & \textbf{0.13} & 0.45          \\
CFARnet & \textbf{0.38} & \textbf{0.87} & \textbf{0.85} & 0.08          & \textbf{0.67} \\
\bottomrule

\end{tabular}
\end{sc}
% \end{scriptsize}
\end{center}
\vskip -0.1in
\label{table:corrlated wc}

\end{table}

\bibliographystyle{plain}

\bibliography{main.bib}

% There are two important things that are needed to be notices:
% \begin{itemize}
%     \item A test is defined by both the detector $T(\x)$ and the threshold $\gamma$. Two test that have the same detector but not the same threshold are not equivalent.
%     \item Two tests that are equal up to a \textbf{known} monotonic function that is applied on both the detector and the threshold are equivalent.
% \end{itemize}
\end{document}